\documentclass[a4paper]{jpconf}
\usepackage{graphicx} 
\usepackage{booktabs} 
\usepackage{amsmath} 
\usepackage{multirow} 
\usepackage{ragged2e} 
\usepackage{amsfonts} 
\usepackage[parfill]{parskip}

\begin{document}
\title{Training and Classification using a Restricted Boltzmann Machine on the D-Wave 2000Q}

\author{Vivek Dixit$^{1,3}$, Raja Selvarajan$^{1,3}$, Muhammad A.Alam$^{2,4}$, Travis S. Humble$^5$, and *Sabre Kais$^{1,3,4}$}

\address{$^1$ Department of Chemistry, Purdue University, West Lafayette, IN\\
$^2$ Department of Electrical and Computer Engineering , Purdue University, West Lafayette, IN\\
$^3$ Department of Physics and Astronomy, Purdue University, West Lafayette, IN\\
$^4$ Birck Nanotechnology Center, Purdue University, West Lafayette, IN\\
$^5$ Quantum Computing Institute, Oak Ridge National Laboratory, Oak Ridge, TN}

\ead{kais@purdue.edu}

\begin{abstract}
Restricted Boltzmann Machine (RBM) is an energy based, undirected graphical model.  It is commonly used for unsupervised and supervised machine learning. Typically, RBM is trained using contrastive divergence (CD). However, training with CD is slow and does not estimate exact gradient of  log-likelihood cost function. In this work, the  model expectation of  gradient learning for RBM has been calculated using a quantum annealer (D-Wave 2000Q), which is much faster than  Markov chain Monte Carlo (MCMC)  used in CD. Training and classification results are compared with CD.  The classification accuracy results  indicate similar performance of both methods. Image reconstruction as well as log-likelihood calculations are used to compare the performance of quantum and classical algorithms for RBM training.  It is shown that the samples obtained from quantum annealer can be used to train a RBM on a 64-bit  `bars and stripes' data set with classification performance similar to a RBM trained with CD. Though training based on CD showed improved learning performance, training using a quantum annealer eliminates computationally expensive MCMC steps of CD. 
\end{abstract}

\section{Introduction}

Quantum computing holds promise for a revolution in the field of science, engineering, and industry. Most of the R\&D work related to quantum computing is focused on gate based approach \cite{Rieffel:2014, Neilsen:2002, Kais:2014}, an alternative to this is the adiabatic quantum computing (AQC) \cite{farhi2000quantum, aharonov:2007, Kadowaki:1998, FINNILA1994343}. In AQC, a system of qubits start with a simple Hamiltonian whose ground state is known. Gradually, the initial Hamiltonian evolves into a final Hamiltonian. The final Hamiltonian is designed in such a way that its ground state corresponds to the solution to the problem of interest. According to the quantum adiabatic theorem, a quantum system that begins in the non-degenerate ground state of a time-dependent Hamiltonian will remain in the instantaneous ground state provided the Hamiltonian changes sufficiently slowly  \cite{ Santoro_2006, RevModPhys.80.1061, RevModPhys.90.015002, Boixo2014}. It has been shown theoretically that an AQC machine can solve QMA-complete problems \cite{Biamonte_2008}. QMA-complete is quantum analogue of NP-complete. Thus, AQC can give solutions which are very difficult to find using classical methods.  

  For machine learning and optimization problems, the D-Wave's quantum annealer has been investigated by several  researchers.  Mott \etal. \cite{mott:2017} used D-Wave to classify Higgs-boson-decay signals vs. background. They showed that the quantum annealing-based classifiers perform comparable to the state-of-the-art machine learning methods. Mniszewski \etal. \cite{Ushijima:2017} found that the  results for graph partitioning using D-Wave systems are comparable to commonly used methods. Alexandrov \etal. \cite{Malley:2018} used a D-Wave for matrix factorization. Lidar \etal. \cite{Lidar:2018} used a D-Wave for classification of DNA sequences according to their binding affinities. Kais \etal. have used D-Wave's quantum annealer for prime factorization and electronic structure calculation of molecular systems \cite{Kais2018a, Kais:2018}. 

RBM is a widely used machine learning technique for unsupervised and supervised tasks. However, its training is time consuming due to the calculation of model dependent term in the gradient learning. RBMs are most commonly  trained using a method known as Contrastive Divergence (CD). CD uses Markov chain Monte Carlo (MCMC) which requires long equilibration time. Further, CD does not follow the gradient of  the log-likelihood \cite{hinton:2002}, and is not guaranteed to give correct results. Therefore, better sampling methods can have a positive impact on RBM learning. In this regard, several  researchers have investigated the D-Wave annealer for RBM applications.  Adachi \etal. \cite{adachi2015application} used quantum annealing for training RBMs, which were further used as layers  of a two layered deep neural network and post-trained by back-propagation algorithm. The authors conclude that the hybrid approach results in faster training, although the relative effectiveness of  RBM trained using a quantum-annealer vs. contrastive divergence has not been documented. Among other works related to the topic, Dumoulin \etal. \cite{Dumoulin} assessed the effect of various parameters like limited connectivity, and noise in weights and biases of RBM on its performance. Koshka \etal.  explored the energy landscape of a RBM embedded onto a D-Wave machine, which was trained with CD \cite{koshka:2018a, koshka:2018b, koshka:2017, koshka:2016}. There has been growing interest in quantum machine learning including Boltzmann machines \cite{Amin:2018, lloyd2013, PhysRevLett.113.130503, wiebe2014quantum}, however training a quantum machine learning models on moderate or large dataset is challenging.

In this work, our objective is to train a RBM using samples from D-Wave quantum annealer and compare its performance with a RBM trained with contrastive divergence.  The model dependent term in the gradient of log-likelihood can be estimated by using samples drawn from a quantum annealer. Trained models were compared based on the classification accuracy, image reconstruction and log-likelihood values. In order to carry out this study,  bars and stripes (BAS) data set has been used. BAS is a binary data set with each record comprising of 64 bits.

\section{Methods}

\subsection{Restricted Boltzmann Machine (RBM)}

A Restricted Boltzmann Machine is an energy based model, inspired from Boltzmann distribution of energies of Ising model of spins. It models the underlying probability distribution of data set and can be used for machine learning applications. A RBM is comprised of two layers of binary variables  known as visible and hidden layers. The variables or units in visible and hidden layers are denoted as $\{ v_1, v_2, ....,v_N\}$ and $\{ h_1, h_2, ....,h_M\}$, respectively. The variables in one layer interact with the variables in the other layer,  however interactions between the variables in the same layer are not permitted. The energy of the model is given by:
\begin{equation}\label{eq:1}
E = - \sum_i b_i v_i - \sum_j c_j h_j - \sum_{i,j} v_i w_{ij}h_j,
\end{equation}
where $b_i$ and $c_j$ are bias terms; $w_{ij}$ represents the strength of interaction between variables $v_i$ and $h_j$. Let us represent the variables in the visible layer collectively by a vector: $v \in \{ 0,1\}^N$, similarly for the hidden layer: $h \in \{ 0,1\}^M$. Using this representation equation (\ref{eq:1}) can be written as:
\begin{equation}\label{eq:2}
E(v,h)=-b^Tv-c^Th-h^TWv,
\end{equation}
where $b$ and $c$ are bias vectors at the visible and hidden layer, respectively; $W$ is a weight matrix composed of $w_{ij}$ elements. The probability that the model assigns to the configuration $\{v, h\}$ is:

\begin{equation}\label{eq:3}
P(v,h)=\frac{1}{Z} e^{-E(v,h)}, \qquad Z=\sum_v \sum_h e^{-E(v,h)},
\end{equation}

where $Z$ is partition function. Substituting value of $E(v,h)$, from equation  (\ref{eq:2}), we get:

\begin{equation}\label{eq:4}
Z=\sum_v \sum_h e^{b^Tv+c^Th+h^T\cdot W\cdot v} = \sum_h e^{c^Th} \sum_v e^{b^Tv + h^T\cdot W\cdot v}
\end{equation}

\begin{equation}\label{eq:5}
Z= \sum_h e^{c^Th} \sum_v e^{(b^T + h^T\cdot W)v} =  \sum_h e^{c^Th} \sum_v e^{s\cdot v},
\end{equation}
where $s$ is:
\begin{equation}\label{eq:6}
s = b^T + h^TW = [s_1, s_2, . . .,s_N];
\end{equation}

N is the number of variables in the visible layer. Now, $Z$ can be written as:

\begin{equation}\label{eq:7}
Z= \sum_{h} e^{c^Th} \prod_{j=1}^N( 1+ e^{s_j})
\end{equation}

From equation (\ref{eq:7}), we notice that the calculation of $Z$ involves summation over $2^M$ configuration, where $M$ is the number of variables in hidden layer. On the contrary, we need $2^{M+N}$ configurations to evaluate $Z$ using equation (\ref{eq:3}).

\subsection{Minimization of the log-likelihood}

The partition function $Z$ is hard to evaluate. The joint probability, $P(v,h)$, being a function of $Z$ is also hard. Due to the bipartite graph structure of the RBM, the conditional distributions $P(h|v)$ and $P(v|h)$ are simple to compute,

\begin{equation}\label{eq:8}
P(h|v)= \frac{P(v,h)}{P(v)}
\end{equation}
where $P(v)$ is given by the following expression:
\begin{equation}\label{eq:9}
P(v)= \frac{\sum_h e^{-E(v,h)}}{Z}.
\end{equation}
Substituting values from equation  (\ref{eq:3}) and equation  (\ref{eq:9}) into equation  (\ref{eq:8}) gives:
\begin{equation}\label{eq:10}
P(h|v)= \frac{exp\{{\sum_j c_j h_j + \sum_j (v^T W)_j h_j}\} }{Z^\prime},
\end{equation}

where
\begin{equation}\label{eq:11}
 Z^\prime = \sum_h exp(c^Th + h^TWv).
\end{equation}

\begin{equation}\label{eq:12}
P(h|v)= \frac{1}{Z^\prime} \displaystyle\prod_{j} exp\{{c_j h_j + (v^T W)_j h_j}\}  
\end{equation}

Let's denote

\begin{equation}\label{eq:13}
\widetilde{P}(h_j|v) = exp\{c_j h_j + (v^T W)_j h_j\}
\end{equation}

Now, the probability to find an individual variable in the hidden layer $h_j = 1 $  is:
\begin{equation}\label{eq:14}
P(h_j = 1|v) =\frac{\widetilde{P}(h_j=1|v)}{\widetilde{P}(h_j=0|v) + \widetilde{P}(h_j=1|v)} = \frac{exp\{c_j + (v^T W)_j \}}{1+exp\{c_j + (v^T W)_j \}}
\end{equation}

Thus, the individual hidden activation probability is given by:

\begin{equation}\label{eq:15}
P(h_j = 1|v) = \sigma\Big(c_j + (v^T W)_j \Big),
\end{equation}

where $\sigma$ is the sigmoid function. Similarly, the activation probability of a visible variable conditioned on a hidden vector $h$  is given by:

\begin{equation}\label{eq:16}
P(v_i = 1|h) = \sigma\Big(b_i + (h^T W)_i \Big).
\end{equation}

A RBM is trained by maximizing the likelihood of the training data. The log-likehood is given by:

\begin{equation}\label{eq:17}
l(W,b,c)= \sum_{t=1}^{N} \log P\Big(v^{(t)}\Big) = \sum_{t=1}^{N} \log \sum_{h} P\Big(v^{(t)},h\Big)
\end{equation}

\begin{equation}\label{eq:18}
l(W,b,c)= \sum_{t=1}^{N} \log \sum_{h} e^{-E(v^{(t)},h)} - N \cdot \log \sum_{v,h} e^{-E(v,h)}.
\end{equation}

where $v^{(t)}$ is a sample from the training dataset. Denote $ \theta=\{W, b, c\} $. The gradient of the log-likelihood is given by:

\begin{equation}\label{eq:19}
\nabla_{\theta} l(\theta) = \sum_{t=1}^{N} \frac{\sum_h e^{-E(v^{(t)},h)} \nabla_\theta (-E(v^{(t)},h))}{\sum_h e^{-E(v^{(t)},h)}}  - N \cdot \frac{\sum_{v,h} e^{-E(v,h)} \nabla_\theta (-E(v,h))}{\sum_{v,h} e^{-E(v,h)}} 
\end{equation}

\begin{equation}\label{eq:20}
\nabla_{\theta} l(\theta) = \sum_{t=1}^{N} \langle\nabla_\theta(-E(v^{(t)},h))\rangle_{P(h \mid v^{(t)})} - N \cdot \langle\nabla_\theta(-E(v,h))\rangle_{P(v, h)},
\end{equation}

where $\langle \cdot \rangle_{P(v, h)}$ is expectation value with respect to distribution $P(v,h)$. Since $ \theta=\{W, b, c\} $, therefore the gradient with respect to $\theta$ can also be expressed in terms of its components:

\begin{equation}\label{eq:21}
\nabla_{w} l = \frac{1}{N}\sum_{t=1}^{N} \langle v^{(t)}\cdot h^{(t)}\rangle_{P(h \mid v^{(t)})} -  \langle v\cdot h\rangle_{P(v, h)}
\end{equation}

\begin{equation}\label{eq:22}
\nabla_{b} l = \frac{1}{N}\sum_{t=1}^{N} \langle v^{(t)}\rangle_{P(h \mid v^{(t)})} -  \langle v\rangle_{P(v, h)}
\end{equation}

\begin{equation}\label{eq:23}
\nabla_{c} l = \frac{1}{N}\sum_{t=1}^{N} \langle h^{(t)}\rangle_{P(h \mid v^{(t)})} -  \langle h\rangle_{P(v, h)}
\end{equation}

The first term in equation (\ref{eq:20}) is expectation value of $\nabla_\theta(-E(v^{(t)},h))$ with respect to Boltzmann distribution, $v^{(t)}$ is a row vector from the training dataset and $h$ is a hidden vector. Given $v^{(t)}$, $h$ can be calculated via equation (\ref{eq:15}). 

The second term in equation  (\ref{eq:20}) is a model dependent term, expectation value of $\nabla_\theta(-E(v,h))$, $v$ and $h$ can be any possible binary vectors. This term is difficult to evaluate as it requires all possible combinations of $v$ and $h$. Generally this term is estimated using  contrastive divergence, where one uses many cycles of Gibbs sampling to transform the training data into data drawn from the proposed distribution. We used equation  (\ref{eq:15}) and  (\ref{eq:16}) to sample from hidden and visible layers repeatedly. Once we have the gradient of  log-likelihood (equation (\ref{eq:18})), weights and biases can be estimated using gradient accent optimization:

\begin{equation}\label{eq:24}
  \theta_j^{new} = \theta_j^{old} + \epsilon \cdot \nabla_{\theta_j} l(\theta_j)
\end{equation}
where $\epsilon$ is the learning rate.  

 \textit{Alternatively, the second term can be calculated using samples drawn from the D-Wave quantum annealer, which is a much faster procedure than MCMC}.

\begin{figure}
\begin{center}
\includegraphics[scale=0.7]{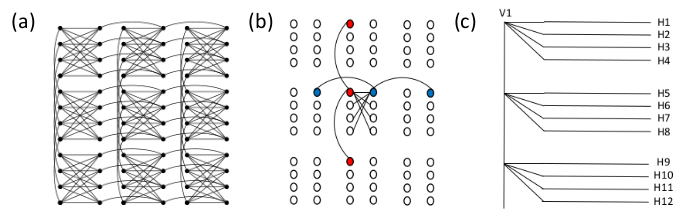}
\end{center}
\caption{\label{label2} (a) C3 Chimera graph of qubits. (b) Three vertical (red) and horizontal (blue) qubits are chained to form a visible and a hidden unit. (c) Connectivity of visible unit (V1) with 12 hidden units. Here, each unit is formed by ferromagnetic couplings between three qubits.} 
\end{figure}

\subsection{D-Wave Hamiltonian and arrangement of qubits}  
The Hamiltonian for a D-Wave system of qubits can be represented as: 

\begin{equation}\label{eq:25}
h_{Ising}= -\frac{A(s)}{2}\Big(\sum_{i}\hat{\sigma}_x^{i}\Big)+\frac{B(s)}{2}\Big( \sum_{i} h_i \hat{\sigma}_{z}^{i} + \sum_{(i>j)} J_{ij} \hat{\sigma}_z^{i} \hat{\sigma}_z^{j}\Big)
\end{equation}

where $\hat{\sigma}_{x,z}^{(i)}$ are Pauli matrices operating on qubit $q_i$.  $h_i$ and $J_{i,j}$ are the qubit biases and coupling strengths. $s$ is called the anneal fraction. $A(s)$ and $B(s)$ are known as anneal functions. At $s=0$, $A(s) \gg B(s)$, while  $A(s) \ll B(s)$ for $s=1$. As we increase $s$ from 0 to 1, anneal functions change gradually to meet these boundary conditions. In the standard quantum annealing protocol, $s$ changes from 0 to 1. The network of qubits starts in a global superposition over all possible classical states and after $s = 1$, the system is measured in a single classical state. 

The arrangement of qubits on the D-Wave chip forms a C16 Chimera graph with $16 \times 16$  unit cells (2048 qubits are mapped into a $16 \times 16$ matrix  of unit cells; each unit cell has 8 qubits). Figure~\ref{label2}(a)  shows a C3 Chimera graph with $3 \times 3$ unit cells. Within each unit cell there are two sets of 4 qubits which are connected to each other in a bipartite fashion. As shown in the figure, each qubit in a unit cell is  connected to four qubits of the unit cell and two qubits of other unit cells. Thus, each qubit can be connected to a maximum of 6 qubits. This connectivity can be enhanced by forming a strong ferromagnetic couplings between the qubits, which forces coupled qubits to stay in the same state.

\subsection{RBM embedding onto the D-Wave QPU}  
Mapping an AQC algorithm on a specific hardware is nontrivial and requires creative mapping. There are several algorithms that can be used to map a graph to the physical qubits on adiabatic quantum computer\cite{Humble:2018, Date:2019}. However, it is nontrivial to find a simple embedding when the graph size is large. Taking into consideration the arrangement of qubits on 2000Q processor, we found a simple embedding which utilizes most of the working qubits.   In the present study, we used RBMs in two configurations, one with 64 visible unit and 64 hidden units, another with 64 visible units and 20 hidden units. Here, we will discuss embedding of the RBM with 64 units in both layers. 

Each unit of the RBM is connected to 64 other units, but in the D-Wave each qubit only connects to 6 other qubits. To enhance the connectivity qubits can be coupled together or cloned by setting $J_{i,j}=-1$ . This forces the two qubits to stay in the same state.  In our embedding one unit of  RBM is formed by connecting 16 qubits. The D-Wave processor has qubits arranged in  $16 \times 16$ matrix  of unit cells. Each unit cell has two sets of 4 qubits arranged in bipartite fashion. Each qubit in the left column of the unit cell can be connected to one qubit of the unit cell just above it and one just below it. There are 16 unit cell along one side, so a chain of 16 qubits can be formed. This chain represents one visible unit of the RBM.  Figure~\ref{label2}(b) shows procedure to couple 3 qubits to form a chain which represents a visible unit for a $3 \times 3$ chimera graph. Qubits that form one visible unit are shown in red forming a vertical chain. Since, there are 4 qubits in left column of the unit cell, four chains can be formed resulting in 4 visible units of RBM.  Four qubits that form the right column of the unit cell can be connected to form horizontal chains, as shown in Figure~\ref{label2}(b). These horizontal chains form the hidden units. There are 16 unit cells along the horizontal direction, therefore each horizontal chain is  also composed of 16 qubits. Exploiting the arrangement of qubits of the D-Wave QPU, 64 vertical and 64 horizontal chains can be formed representing the 64 visible and 64 hidden units of RBM.  Figure~\ref{label2}(c) shows the scheme that we used to connect one visible unit (V1) to 12 hidden units. In this fashion one can embed a RBM with 64 visible and 64 hidden units on a C16 Chimera graph. Of course care must be taken for any inaccessible qubits to form a further restricted RBM. In our experiments, we found that absence of few qubits does not affect the performance of the resulting network. Employing this embedding we were able to connect most of the RBM units to 64 units of the opposite layer. 

\subsection{Classification and image reconstruction}

Each record of the bar and stripes (BAS) data set has 64 bits. The last two bits are for labeling the pattern: 01 for a bar and 10 for a stripes pattern ( Figure~\ref{label2}(a)). If the last two bits are 00 or 11, prediction by RBM is incorrect. Once we obtained the weights and biases of the RBM from the training step, RBM can be used for classification or image reconstruction. To predict the class of a test record we apply its first 62 bits  at  the visible layer ( Figure~\ref{label2}(b)). For the last two classifying bits (L1 and L2),  we randomly input either zero or one. We then run 50 Gibbs cycles, keeping the 62 visible units clamped at the values of the test record. At the end of 50 Gibbs cycles label units 63 and 64 are read. $L1=0$  and $L2=1$ indicates a bar pattern, while $L1=1$  and $L2=0$ suggests a stripe pattern.
For the problem of image reconstruction, the goal is to predict value of missing part of the image. Similar procedure can applied for image reconstruction where a trained RBM is used to predict values of the missing units, while in the classification process one  only predicts the classifying labels. In this case we clamp the visible units where value are given, run few Gibbs cycle and then sample the units where values are missing.

A different procedure is used for classification and image reconstruction using a quantum annealer. First step is to embed a trained RBM onto the D-Wave QPU. Then a field `h' which is proportional to the test record (or corrupted image) is applied to the visible units. Finally, quantum annealing step is performed which results in predicting values of missing visible units.

\begin{figure}
\begin{center}
\includegraphics[scale=1.0]{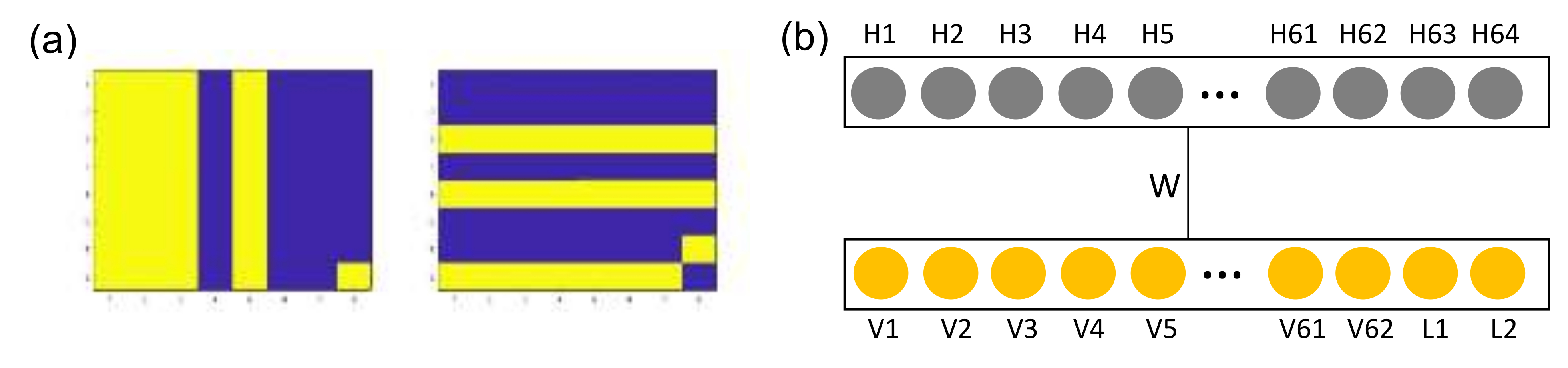}
\end{center}
\caption{\label{records_rbm} (a) Example of bar (left) and stripe (right) pattern of size  $8 \times 8$. A blue cell represents a 0 while a yellow cell a 0.  Last two bits (bottom right) are labels; bar = 01, stripe = 10. (b) RBM for classification. Box with yellow units is the visible layer. Hidden layer is shown by box with grey units. A record of size $1\times64$ is applied to the visible layer. Units L1 and L2 are the labels.} 
\end{figure}

\section{Results and Discussion}

\begin{figure}
\begin{center}
\includegraphics[scale=1.0]{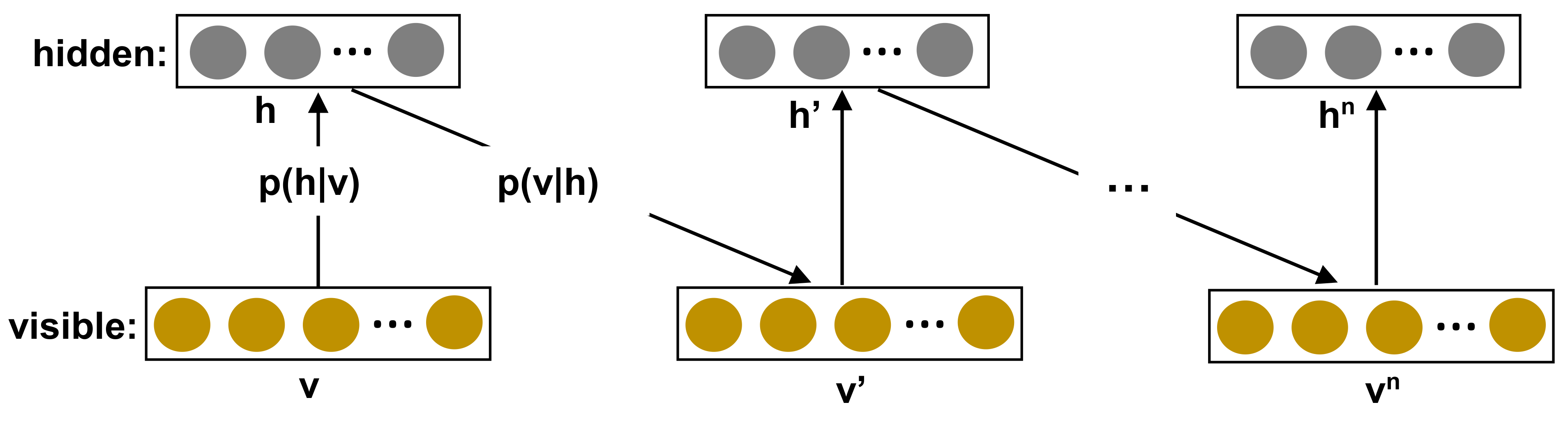}
\end{center}
\caption{\label{fig:cd-n} Illustration of the contrastive divergence algorithm for training restricted Boltzmann machines (RBMs). If sampling stops at $n^{th}$ Gibbs step of the Markov Chain, the procedure is known as CD-n.  
} 
\end{figure}
In the present work, we have used bars and stripes (BAS) data set, which is widely used as a benchmark for training and classification using RBMs. BAS is a binary data set consisting of records 64 bit in length, with the last 2 bits representing the label of the record: 01 for a bar and 10 for a stripe pattern. Our  BAS data set is comprised of 500 unique records. The number of unique samples used for training is 300, with the remaining 200 samples were held for testing. Classification of bars and stripes, image reconstruction and log-likelihood were used to compare the performance of trained RBMs. 

\subsection{RBM Training}

Equation (\ref{eq:18}) has been used to train the RBMs.  The first term in this equation is data dependent. This term can be exactly calculated using  the conditional probabilities $P(h|v)$  and $P(v|h)$ given by equation (\ref{eq:13}) and (\ref{eq:14}).  The second term  is  model dependent, which requires expectation value over all possible hidden $h$ and visible $v$ vectors, which is clearly intractable. Typically, the model dependent term is approximately estimated using a method known as contrastive divergence (CD). In this approach, samples needed to calculate the model dependent term are obtained by running Gibbs chain starting from a sample from the training data (figure \ref{fig:cd-n}).  If n Gibbs steps are performed, method is known as CD-n. It is shown by Hinton that $n=1$ could be sufficient for convergence (CD-1) \cite{Hinton2002}. In CD-1, first a data sample is applied at the visible layer, then  equation (\ref{eq:13}) is used to generate corresponding hidden vector at the hidden layer. Now, this hidden vector is used to generate a new visible vector using equation (\ref{eq:14}), which is in turn used to generate a new hidden vector. These new visible and hidden vectors are used to calculate model expectation value. This process is repeated for each record in the data set.  A detailed description of  RBM training using CD  is given in a review article by Hinton \cite{Hinton:2012}. The model dependent term can also be calculated using samples ($v$ and $h$) obtained from a RBM mapped on the D-Wave.   From equation (\ref{eq:17}) we notice that in the second term expectation value should be calculated with respect to $e^{-E(v,h)}$ distribution, while samples from the D-Wave follow a distribution of $e^{-\frac{E(v,h)}{kT}}$.  Following the approach used by Adachi \etal. \cite{adachi2015application}, we used a empirical parameter, $S$, such that for the model dependent term, we sample from  $e^\frac{-E(v,h)}{SkT}$ distribution. Here, $S$ is  a parameter, which is determined by trial and error method.  The optimal condition corresponds to the case when $SkT=1$. A different approach was taken by Benedetti \etal. \cite{PhysRevA.94.022308}. They calculated effective temperature during each epoch.  Their approach is difficult to apply in the present case of 64 bits record length BAS data set. BAS data set that they used had 16-bit records. A complex data set leads to a complicated distribution, which makes training with a variable temperature difficult.  

\begin{figure}
\begin{center}
\includegraphics[scale=0.5]{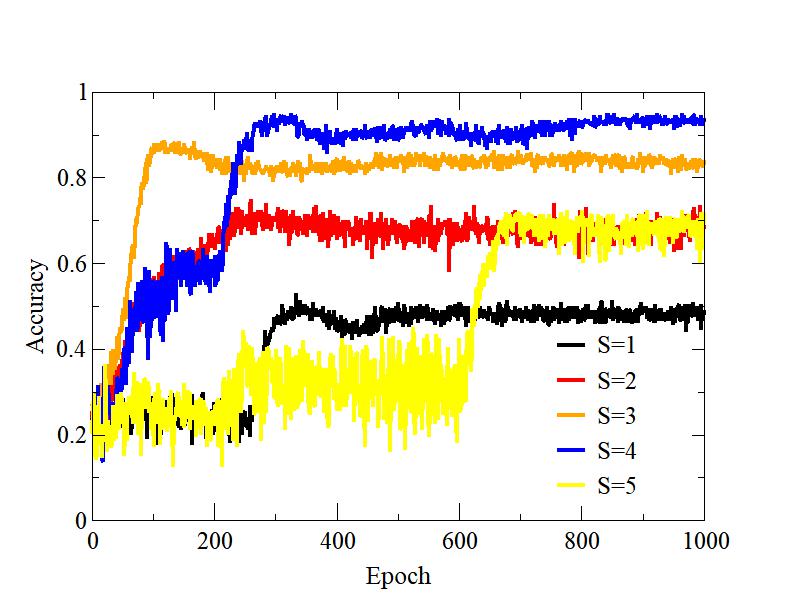}
\end{center}
\caption{\label{fig:scale7} Classification accuracy vs. epoch graph on BAS dataset. The plot shows the effect of empirical parameter $S$ on the accuracy. The best results were obtained for $S=4$.  
} 
\end{figure}

In order to train a RBM using D-Wave,  model parameters ($w_{ij}$, $b_i$ and $c_j$) were initialized with random values, first term of equation (\ref{eq:18}) was calculated exactly using these weights and biases and the training dataset. Weights and biases were used to embed the RBM onto the D-Wave QPU, and quantum annealing was performed. Once annealing was complete, the D-Wave returned low energy solutions. Based on the mapping of the RBM, visible $v$ and hidden $h$ vectors were obtained from the solutions returned from the D-Wave. These $v$ and $h$ samples were used to calculate the model dependent expectation value which in turn gives the gradient of  log-likelihood.  The gradient was further used to calculate new weights and biases. The whole process was repeated until some convergence criterion was achieved. An optimal value of the empirical parameter $S$ is  important for the correct sampling of $v$ and $h$ vectors. The effect of change in $S$ on the classification accuracy is shown in figure (\ref{fig:scale7}), a plot between accuracy and epoch. The term epoch means a full cycle of iterations, with each training pattern participating only once. Accuracy is defined as:

\begin{equation}\label{eq:26}
\text{Accuracy} = \frac{\text{Number of correct predictions}}{\text{Total number of predictions}}
\end{equation}

 The classification accuracy is maximum for $S=4$. The performance of the model during the training process can be visualized by plotting classification accuracy with epochs. Figure (\ref{fig:accuracy}) shows the plot of classification accuracy vs. epochs for bars (left) and stripes (right) patterns. This calculation was performed on the test data set.  As the number of epoch increases from 0 to 400, the classification accuracy increases after that it stays constant. Based on these results, we conclude that performance of RBM trained with quantum annealing is similar to that using CD-1. However, the classification accuracy from CD-1 based training shows higher fluctuations. 

\subsection{Image reconstruction}

\begin{figure}
\begin{center}
\includegraphics[scale=0.37]{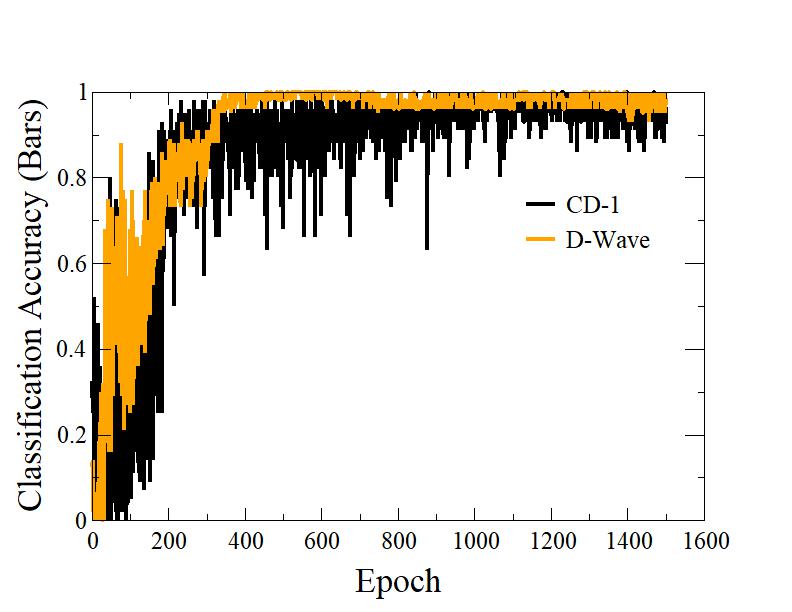}
\includegraphics[scale=0.37]{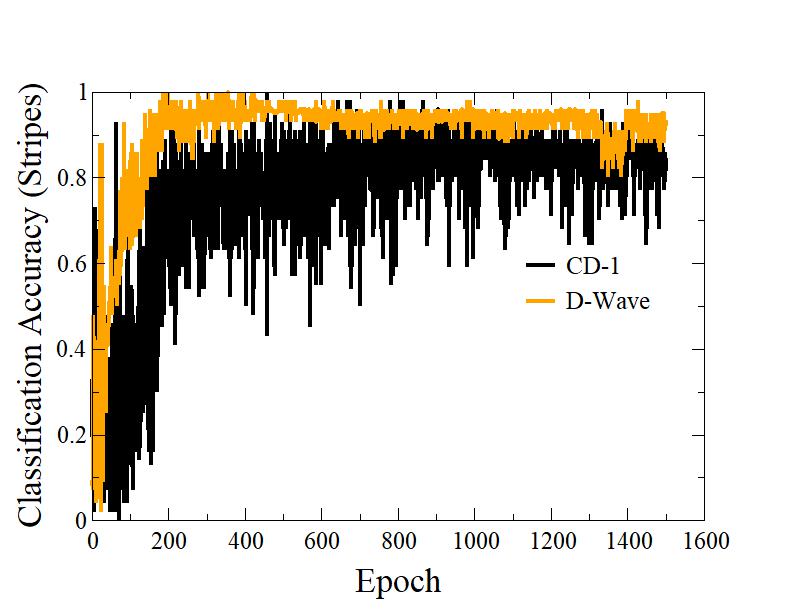}
\end{center}
\caption{\label{fig:accuracy} Plots showing classification accuracy of individual classes with epoch. Gradient of log-likelihood was calculated using samples generated via contrastive divergence and D-Wave's quantum annealing. Comparison of accuracy of both methods are presented for bars (left) and stripes (right) patterns. 
} 
\end{figure}

\begin{figure}
\begin{center}
\includegraphics[scale=1.0]{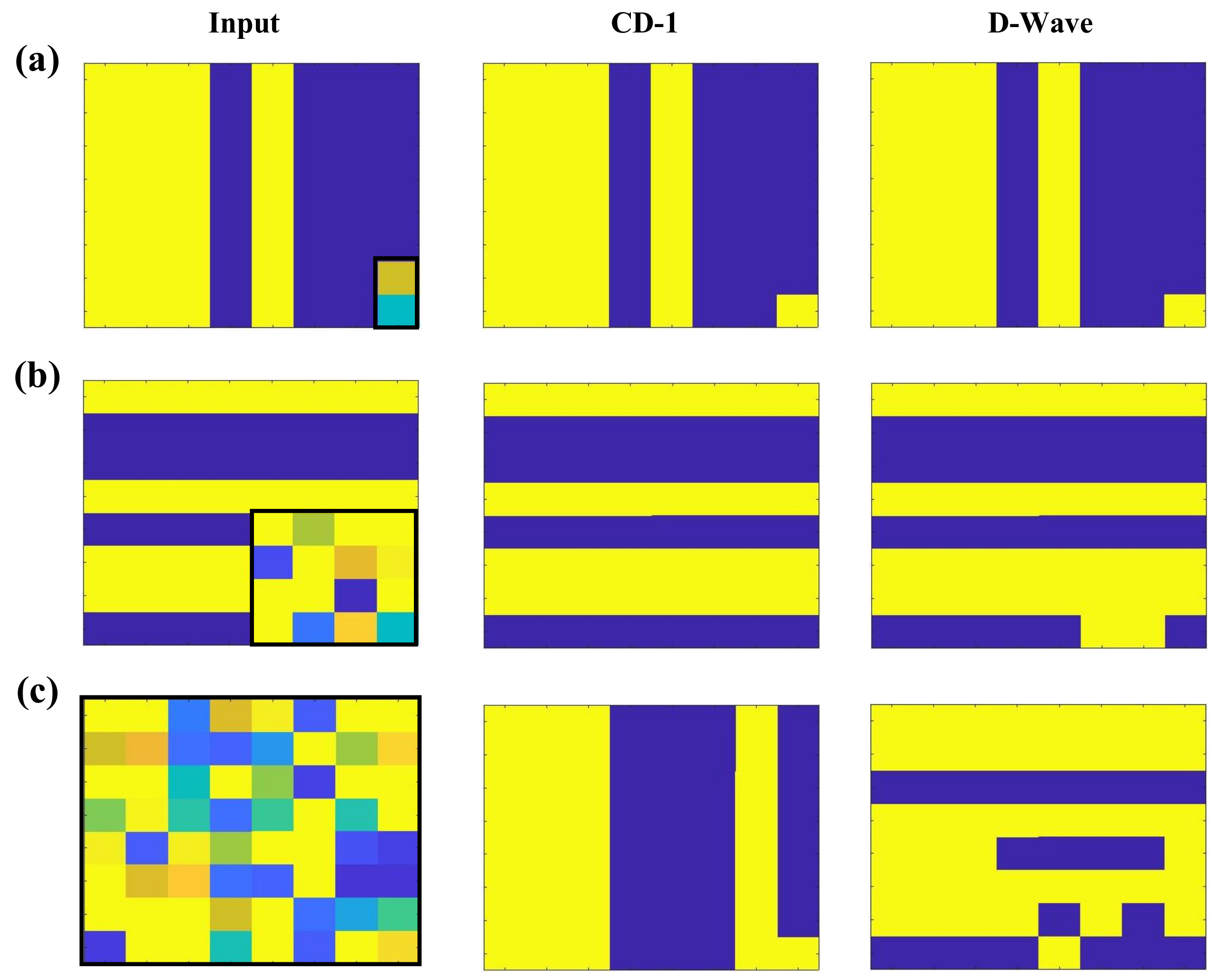}
\end{center}
\caption{\label{fig:reconstruct}
Image reconstruction. Original data was first corrupted, then reconstructed. The images in the left column are data fed to the RBMs; output from the RBMs trained using CD-1 and quantum annealing (D-Wave) are shown in the middle and right column, respectively. Corrupted bits in the input images are enclosed by black rectangles.} 
\end{figure}

 For classification task both training methods (quantum annealing and CD-1) showed similar results. Classification task requires prediction of target labels based on the features in the data set. Input data record is applied at the visible layer and  the target labels are reconstructed. More difficult task would be the reconstruction of not just the target labels, but also some other bits of the record. We call this task -  image reconstruction. Here, we take a 64 bit record from the test data set, corrupt some of its bits, and then apply this modified test record to the visible layer of a trained RBM. We follow the procedure explained earlier for image reconstruction using RBM trained with CD-1 and quantum annealing. The results of  image reconstruction are presented in figure (\ref{fig:reconstruct}). In figure \ref{fig:reconstruct}(a) only target labels were corrupted/reconstructed. We notice that both training methods correctly reproduced the classifying labels. In the second case, figure \ref{fig:reconstruct}(b), 16 bits of the  original data record were corrupted. The RBM trained using CD-1 correctly predicted all the bit, while two bits were incorrectly predicted by the RBM trained with quantum annealing. In the third case, figure \ref{fig:reconstruct}(c), a completely random 64 bit  input vector (all bits corrupted) was fed to the both RBMs. In the case of CD-1, the output is a bar pattern, where as the  D-Wave trained RBM resulted in a stripes pattern with many bits incorrectly predicted.  From these results, we infer that the RBM trained using CD-1 performs slightly  better than the RBM trained using quantum annealing. The first example of image reconstruction (figure \ref{fig:reconstruct}(a)) where only target labels were predicted and the plot of classification accuracy vs epochs (figure (\ref{fig:accuracy}))  indicate that it is easier to predict the classifying labels of a record compared to the prediction of other bits.  Similar results were obtained (not shown here) when other records of the test dataset were investigated for image reconstructions.

\subsection{Log-likelihood comparison}

\begin{figure}
\begin{center}
\includegraphics[scale=0.5]{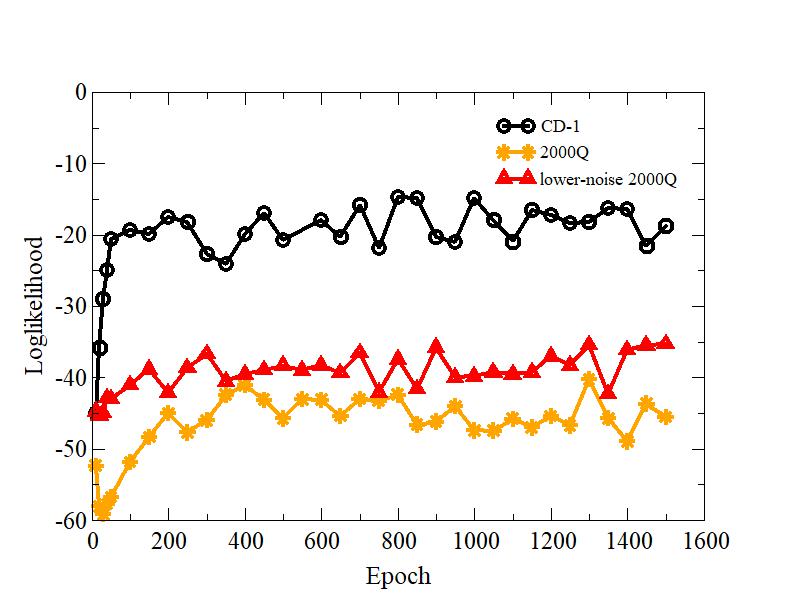}
\end{center}
\caption{\label{fig:logL} Plot showing variation in log-likelihood with training epochs. Label CD-1 represents RBM training employing contrastive divergence, while label 2000Q indicates training using samples are generated from D-Wave. The lower-noise 2000Q processor results in slightly improved log-likelihood} 
\end{figure}

The classification accuracy results indicated similar performances of both methods  (CD-1 and quantum annealing). However, image reconstruction suggests improved performance of  CD-1.  In order to further compare and quantify the performances of these two models log-likelihood of training data was calculated. Several researchers have used  `log-likelihood' in order to compare different RBM models \cite{Tieleman:2008, Cho:2010}. The log-likelihood has been computed using equation (\ref{eq:16}). It involves the computation of  partition function $Z$. If number of units in the hidden layer is not too large, $Z$ can be exactly calculated using equation (\ref{eq:7}).  To calculate the log-likelihood, the number of hidden units was set to 20. The log-likelihood of both models were computed at various epochs. The results are presented in Figure \ref{fig:logL}. We notice that the log-likelihood is higher for  RBM trained using CD-1 compared to quantum annealing. A lower value of  the log-likelihood for the  D-Wave trained model could be attributed to a restricted range of allowed values for the bias field $h$ and the coupling coefficients $J$. Another reason could be an instance (each set of $h$ and $J$) dependent temperature variation during the RBM training \cite{PhysRevA.94.022308}. This disturbs the learning of embedded RBM. Figure \ref{fig:logL} also compares the log-likelihood values calculated using new lower-noise D-Wave 2000Q processor and an earlier 2000Q processor. D-Wave's  lower-noise machine shows slightly improved log-likelihood values over the entire training range.

\section{Conclusions}

In this work, we presented an embedding that can be use to embed a RBM with 64 visible and 64 hidden units. We trained a RBM by calculating the model dependent term of the log-likelihood using samples obtained from the D-Wave quantum annealer. A trained RBM was embedded onto the D-Wave QPU for classification and image reconstruction.  We also showed that new lower-noise quantum processor gives improved results. The performance of the RBM was compared with a RBM trained with commonly used method called contrastive divergence (CD-1). Though both methods resulted in comparable classification accuracy, CD-1 training resulted in a  better image reconstruction and the log-likelihood values. In our future work, we will improve the RBM training using quantum annealer by finding effective ways to estimate instance dependent temperature of the quantum annealer and by incorporating this temperature in the RBM training procedure. Our present method relies on the estimation of a empirical parameter, which eliminates the need to find D-Wave QPU temperature during each iteration. RBM training using the samples from a quantum annealer removes the need time consuming MCMC steps during training and classification procedures.  These  computationally expensive MCMC steps are essential part of training and classification with CD-1.

\ack{}
We are grateful for the support from Integrated Data Science Initiative Grants (IDSI F.90000303), Purdue University. S. K. would like to acknowledges funding by the U.S. Department of Energy (Office of Basic Energy Sciences) under Award No. DE-SC0019215. This research used resources of the Oak Ridge Leadership Computing Facility, which is a DOE Office of Science User Facility supported under Contract DE-AC05-00OR22725.

\section*{References}
\bibliography{iopart-num}

\end{document}